\documentclass[envcountsame, a4paper]{styles/svproc}
\usepackage{microtype}
\usepackage{graphicx}
\usepackage[font=small]{caption}
\usepackage{subfigure}

\usepackage{booktabs}       
\usepackage{tabularx}
\usepackage{nicefrac}       

\usepackage{amsmath,amsfonts,bm}
\usepackage{amssymb}
\usepackage{mathtools}

\usepackage{pifont}
\usepackage{graphics} 
\usepackage{svg}
\usepackage{tikz}
\usepackage{pgfkeys}
\usepackage{standalone}
\usepackage{cite}
\usepackage{graphicx}
\usepackage{multirow}
\usepackage{threeparttable}
\usepackage{placeins}
\usepackage{adjustbox}
\usepackage[super]{nth}

\usepackage{url}

\usepackage{hyperref}
\usepackage{siunitx}
\DeclareMathOperator*{\argmin}{arg\!\min}

\usepackage{acro}
\newcommand{\newac}[2]{\DeclareAcronym{#1}{short=#1,long=#2}}
\newac{CS}{Constant Strain}
\newac{CoM}{Center of Mass}
\newac{DOF}{Degrees of Freedom}
\newac{EOM}{Equations of Motion}
\newac{GBN}{Generalized Binary Noise}
\newac{HSA}{Handed Shearing Auxetic}
\newac{PCS}{Piecewise Constant Strain}
\newac{RMSE}{Root Mean-Squared Error}

\newcommand{\presub}[2]{\prescript{}{#1}{#2}}

\begin{document}
\mainmatter              
%
\title{An Experimental Study of Model-based Control for Planar Handed Shearing Auxetics Robots}
\titlerunning{Model-based control of planar HSA robots}  
%
\author{Maximilian Stölzle\inst{1} \and Daniela Rus\inst{2} \and Cosimo Della Santina\inst{1}}
%
\authorrunning{Stölzle, Rus, Della Santina} 
%
%
\institute{Delft University of Technology, 2628 CD Delft, Netherlands,\\
\email{M.W.Stolzle@tudelft.nl},\\ 
\and 
Massachusetts Institute of Technology, Cambridge, MA 02139 USA}

\maketitle              

\vspace{-1.5em}

\begin{abstract}
Parallel robots based on Handed Shearing Auxetics (HSAs) can implement complex motions using standard electric motors while maintaining the complete softness of the structure, thanks to specifically designed architected metamaterials.
However, their control is especially challenging due to varying and coupled stiffness, shearing, non-affine terms in the actuation model, and underactuation. In this paper, we present a model-based control strategy for planar HSA robots enabling regulation in task space. 
We formulate equations of motion, show that they admit a collocated form, and design a P-satI-D feedback controller with compensation for elastic and gravitational forces. 
We experimentally identify and verify the proposed control strategy in closed loop.


\end{abstract}
\vspace{-1.5em}\section{Motivation and related work}\vspace{-0.5em}
The deformability, adaptiveness, and compliance of invertebrates serve as an inspiration for continuum soft robots.
While serial continuum soft robots have been intensively investigated in recent years~\cite{della2023model}, parallel soft robots~\cite{hughes2020extensible} 
are less studied despite exhibiting exciting properties such as an improved stiffness-to-weight ratio. 
One recent development in this field is robots based on \acp{HSA}~\cite{truby2021recipe, kaarthik2022motorized, stolzle2024guiding} in which multiple \ac{HSA} rods are connected at their distal end through a rigid platform. 
Twisting of the proximal end of an \ac{HSA} 
causes the rod to elongate 
and enables complex motion primitives in 3D space.
Recent work has investigated 
the mechanical characterization~\cite{good2022expanding}, simulation~\cite{stolzle2023modelling}, and kinematic modeling~\cite{garg2022kinematic, stolzle2023modelling} of \ac{HSA} robots but control has yet to be tackled.
In this work, we make a first step towards achieving task-space control by designing model-based regulators for planar motions. Our approach considers essential characteristics of \ac{HSA} robots, such as underactuation, shear strains, and varying stiffness. 

\begin{figure}[ht]
    \centering
    \subfigure[Blockscheme]{\includegraphics[width=0.62\textwidth]{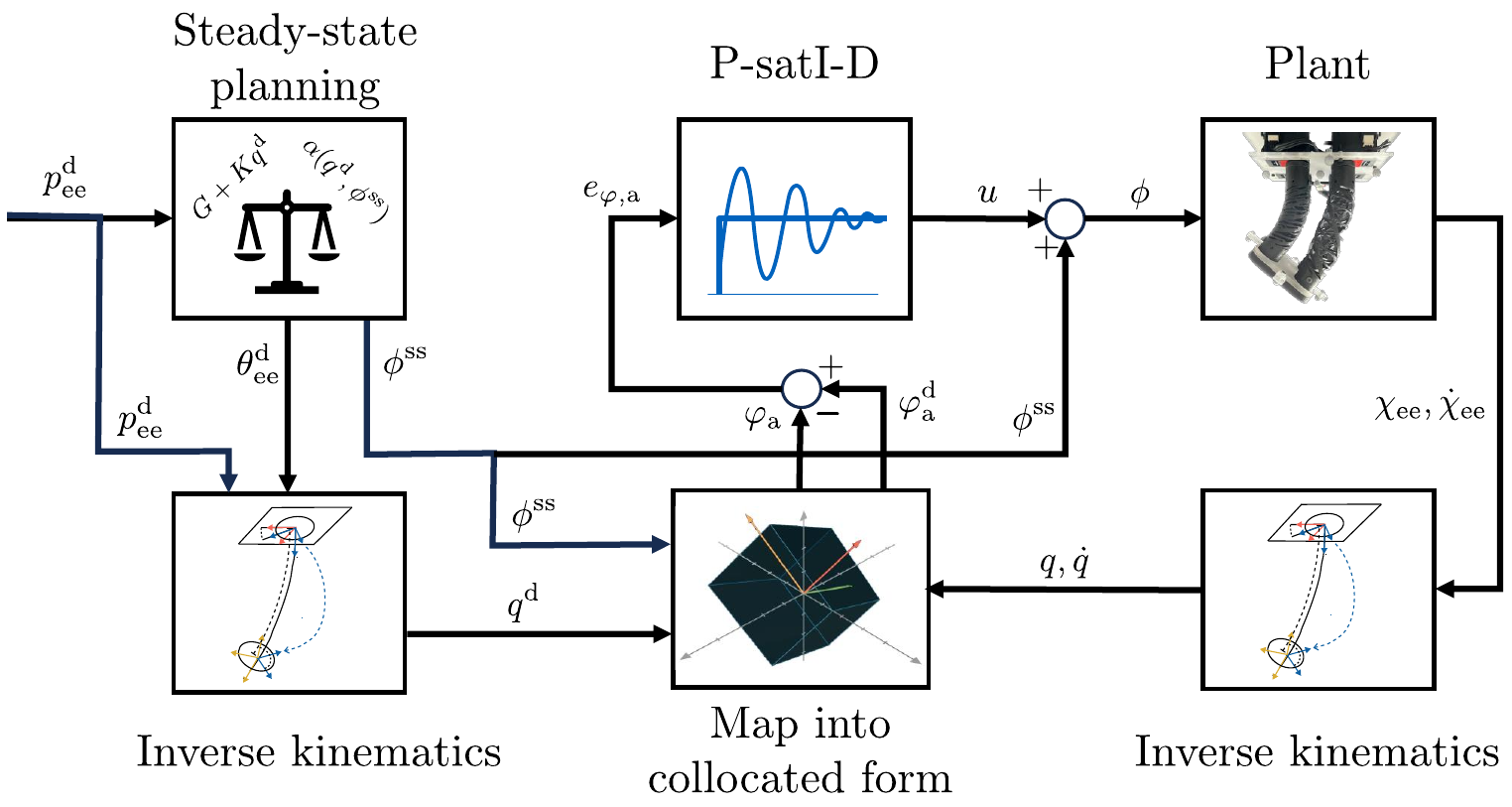}\label{fig:block_scheme_closed_loop_control}}
    \subfigure[Operational workspace]{\includegraphics[width=0.37\columnwidth, trim={7, 7, 7, 7}]{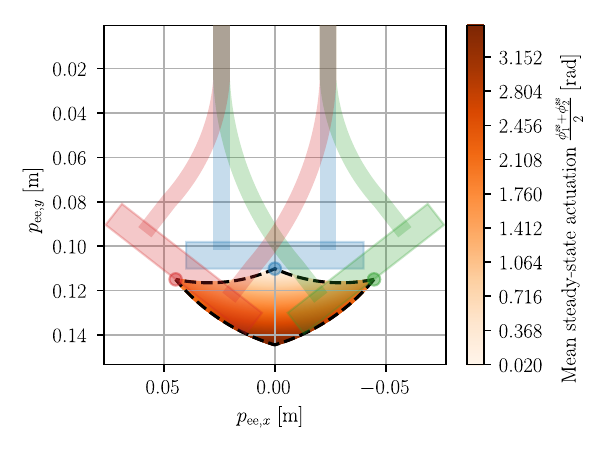}\label{fig:kinematics:workspace}}
    \vspace{-0.2cm}
    \caption{\textbf{Panel (a):} Block scheme of the closed-loop system: we plan the steady-state behavior such that the end-effector matches the given desired position $p_\mathrm{ee}^\mathrm{d}$. The outputs of this planning are the steady-state actuation $\phi^\mathrm{ss}$ and a suitable end-effector orientation $\theta_\mathrm{ee}^\mathrm{d}$. After leveraging inverse kinematics to identify the desired and current configuration, $q$ is mapped into a collocated form where the inputs are decoupled. Finally, we use a P-satI-D feedback controller on the actuation coordinates $\varphi$. \textbf{Panel (b):} Visualization of the operational workspace of a planar HSA robot consisting of FPU rods. The colored area within the black dashed borders represents the positions the end-effector (visualized as a dot) can reach. The coloring denotes the mean magnitude of actuation (i.e., twisting of the rods). Furthermore, we plot three sample configurations: the unactuated straight configuration $q = [0, 0, 0]^\mathrm{T}$ (blue), maximum clockwise bending $q = [\SI{-11.2}{rad \per m}, 0.08, 0.30]^\mathrm{T}$ (red), and maximum counter-clockwise bending $q = [\SI{11.2}{rad \per m}, -0.08, 0.30]^\mathrm{T}$ (green).}
\end{figure}

Kinematic models for parallel robots usually require separate configuration variables for each limb and the enforcement of kinematic constraints~\cite {armanini2021discrete}.
%
We propose to avoid this complexity by 
defining the \ac{CS} of a virtual backbone in the center of the robot to be our configuration variable. 
Subsequently, we derive the system dynamics in Euler-Lagrangian form. We notice that the resulting planar dynamics are underactuated
and that the actuation forces are non-affine with respect to the control inputs, which are the motor angles. The latter is a peculiarity of these systems, rarely observed in other robots.
Based on the model knowledge, we devise a control strategy shown in Fig.~\ref{fig:block_scheme_closed_loop_control} that first maps end-effector positions to desired configurations and steady-state (feedforward) control inputs and then 
also applies a P-satI-D~\cite{pustina2022p} feedback action on the collocated form~\cite{pustina2024input} of the system dynamics.

In summary, we state our contributions as (i) a closed-form solution for the inverse kinematics of a planar \ac{CS} formulation, (ii) an Euler-Lagrangian dynamical model for planar \ac{HSA} robots and its expression in collocated form, (iii) a provably stable model-based control strategy for guiding the end-effector of the robot towards a desired position in Cartesian space, and (iv) experimental verification of both the model and the controller. A video accompanies this paper explaining the methodology and displaying video recordings of the control experiments\footnote{\url{https://youtu.be/7PgKnE_MOsY}}.

%

\vspace{-1em}\section{Technical approach}\vspace{-0.5em}
In the following, we consider a parallel HSA robot moving in a plane. 
First, we derive the kinematic and dynamic models. Subsequently, we devise a planning and control strategy to move the end-effector (i.e., the platform) to a desired position in Cartesian space.

\vspace{-1em}\subsection{Kinematic model}
Following the discrete Cosserat approach~\cite{renda2018discrete}, we characterize the configuration space of the virtual backbone by assuming a \ac{CS} model
$\presub{\mathcal{V}}{\xi}(t) = \begin{bmatrix}\presub{\mathcal{V}}{\kappa}_\mathrm{b} & \presub{\mathcal{V}}{\kappa}_\mathrm{sh} & \presub{\mathcal{V}}{\sigma}_\mathrm{ax}\end{bmatrix}^\mathrm{T} = \mathbb{I}_3 \, q(t) \in \mathbb{R}^3$, where $\kappa_\mathrm{be}$, $\sigma_\mathrm{sh}$, and $\sigma_\mathrm{ax}$ denote the bending, shear, and axial strain respectively.
Given $q$, the pose $\chi = \begin{bmatrix}
    p_x & p_y & \theta
\end{bmatrix}^\mathrm{T} \in SE(2)$, and a point coordinate along the backbone $s \in [0, l^0]$, the forward and inverse kinematics are provided in closed form as
\begin{equation}\small\label{eq:kinematics}
    \chi = \pi(q, s) = \begin{bmatrix}
        \sigma_\mathrm{sh} \, \frac{\mathrm{s}_\mathrm{be}}{\kappa_\mathrm{be}} + \sigma_\mathrm{ax} \, \frac{\mathrm{c}_\mathrm{be}-1}{\kappa_\mathrm{be}}\\
        \sigma_\mathrm{sh} \, \frac{1-\mathrm{c}_\mathrm{be}}{\kappa_\mathrm{be}} + \sigma_\mathrm{ax} \, \frac{\mathrm{s}_\mathrm{be}}{\kappa_\mathrm{be}}\\
        \kappa_\mathrm{be} \, s
    \end{bmatrix},
    \qquad
    q = \varrho(\chi, s) 
    = \frac{\theta}{2s} \: \begin{bmatrix}
        2\\
        p_y - \frac{p_x \, \mathrm{s}_\theta}{\mathrm{c}_\theta-1}\\
        -p_x - \frac{p_y \, \mathrm{s}_\theta}{\mathrm{c}_\theta-1}
   \end{bmatrix},
\end{equation}
where we use the shorthand notations $\mathrm{s}_\mathrm{be} = \sin(\kappa_\mathrm{be}s)$, $\mathrm{c}_\mathrm{be} = \cos(\kappa_\mathrm{be}s)$, $\mathrm{s}_\theta = \sin(\theta)$, and $\mathrm{c}_\theta = \cos(\theta)$.
Furthermore, the forward kinematics of the physical rods $\mathcal{P}_i, \, i \in \{1, 2\}$ can be derived by first following the transformations of the virtual backbone and then adding a local translation $[\pm r_{\mathrm{off}},0]^\mathrm{T}$ with $r_\mathrm{off}$ being the offset distance from the virtual backbone to the centerline of the \ac{HSA} rod. 
After closing the kinematic chain, we identify a mapping $\beta_i: \presub{\mathcal{V}}{\xi} \rightarrow \presub{\mathcal{P}_i}{\xi}$ from the strains of the virtual backbone to the strains in the physical rods: $\beta_i(\presub{\mathcal{V}}{\xi}) = \begin{bmatrix}
    \presub{\mathcal{V}}{\kappa}_\mathrm{b}, & \presub{\mathcal{V}}{\sigma}_\mathrm{sh}, & \presub{\mathcal{V}}{\sigma}_\mathrm{ax} \pm r_{\mathrm{off}}  \presub{\mathcal{V}}{\kappa}_\mathrm{b}
\end{bmatrix}^\mathrm{T}$.
Prior work has shown that the auxetic trajectory of \acp{HSA} can be modeled by coupling the rest length $\tilde{l}_i$ to the twist strain $\kappa_{\mathrm{tw},i}$ of the $i$th \ac{HSA} rod~\cite{stolzle2023modelling, good2022expanding}: $\tilde{l}_i = (1 + \epsilon_i) l^0 = (1 + h_i C_\epsilon \kappa_{\mathrm{tw},i})$ where $l^0$ is the printed length of the rod and $C_\epsilon$ a positive constant.
The handedness $h_i \in \{-1, 1\}$ describes if positive or negative twist angles are needed to elongate the closed \ac{HSA}.
For a given vector of rod twist angles $\phi \in \mathbb{R}^2$ and after defining $\phi_{i}^+ = h_i \phi_i$, the elongation of the $i$th rod is then $\epsilon_i = C_\epsilon \frac{\phi_{i}^+}{l^0}$.
We provide examples in Fig.~\ref{fig:kinematics:workspace} of the operational workspace that can be achieved with this kinematic model.

\begin{figure}[t]
    \centering
    \includegraphics[width=0.85\textwidth]{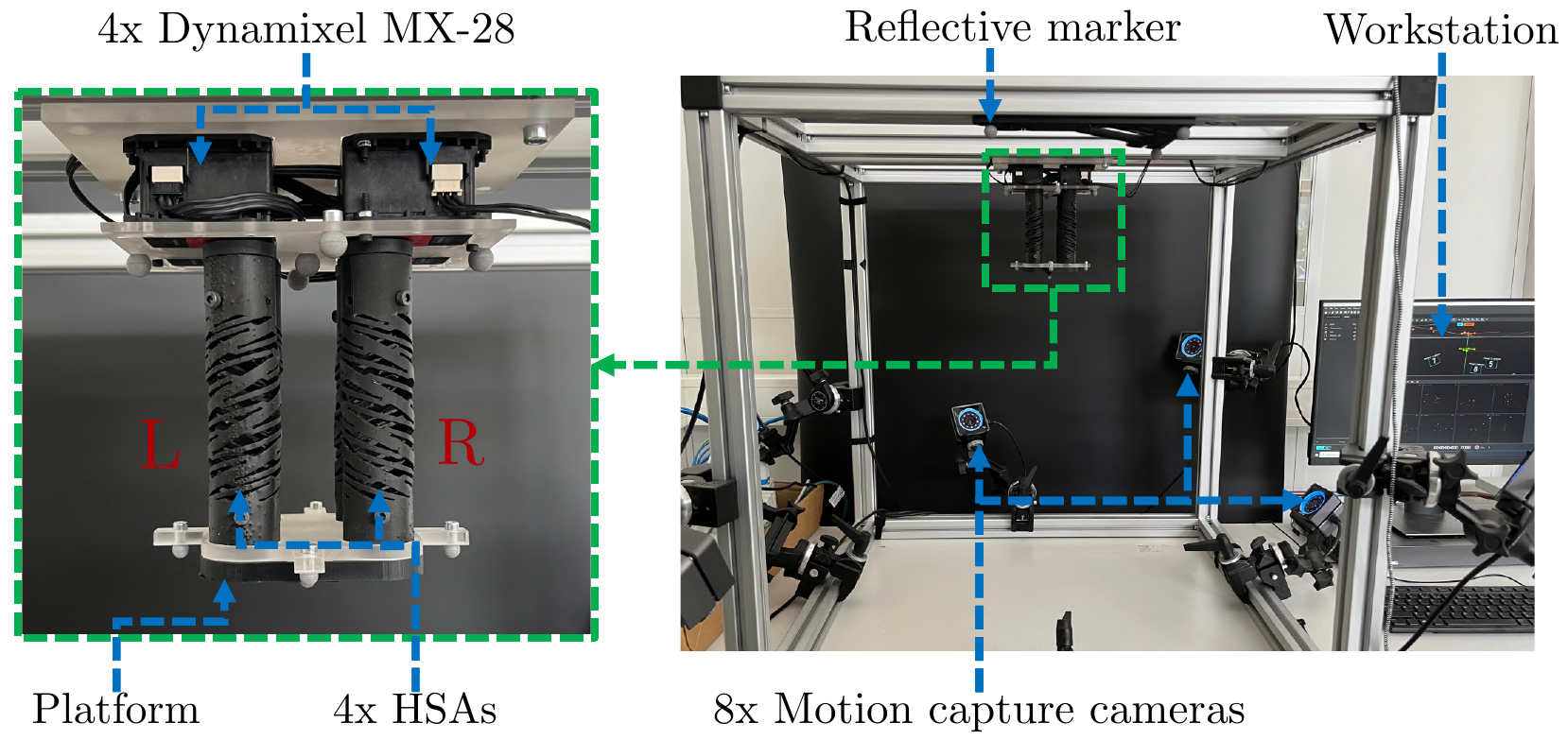}
    \caption{Experimental setup: the parallel robot consists of four HSA rods connected by a platform at their distal end. Four servo motors actuate the HSAs. We track the pose of the end-effector with a motion capture system by attaching reflective markers to the platform.}
    \label{fig:experimental_setup}
\end{figure}

\vspace{-1em}\subsection{Dynamic model}\label{sub:methodology:dynamics}\vspace{-0.4em}
We aim to devise a dynamic model in the Euler-Lagrange form
$M(q) \Ddot{q} + C(q,\dot{q})\dot{q} + G(q) + K (q - q^0) + D \dot{q} = \alpha(q,\phi),$
where $M(q),C(q,\dot{q}),K,D \in \mathbb{R}^{3 \times 3}$ are the inertia, Coriolis (derived with Christoffel symbols), elastic and damping matrices respectively. $q^0 \in \mathbb{R}^3$ captures the rest configuration. The terms $G(q)$ and $\alpha(q,\phi) \in \mathbb{R}^3$ describe the gravitational and actuation forces acting on the generalized coordinates.
The state of the robot at time $t$ can be therefore described by $x(t) = \begin{bmatrix}
    q^\mathrm{T}(t) & \dot{q}^\mathrm{T}
\end{bmatrix}^\mathrm{T} \in \mathbb{R}^6$.
The inertia matrix is found by following the standard procedure of integrating mass and rotational inertia along the \ac{HSA} rods~\cite{della2023model}. Additionally, we consider the inertial contribution of the platform mounted to the distal end of the robot.
Under the small strain assumption, the elastic forces of the $i$th \ac{HSA} rod can be modeled as
\begin{equation}\small
    \presub{\mathcal{P}}{\tau}_{\mathrm{K},i} = 
    \begin{bmatrix}
        S_{\mathrm{be},i}(\phi_i) & S_{\mathrm{b},\mathrm{sh}} & 0\\
        S_{\mathrm{b},\mathrm{sh}} & S_{\mathrm{sh},i}(\phi_i) & 0\\
        0 & 0 & S_{\mathrm{ax},i}(\phi_i)
    \end{bmatrix} 
    \, \left ( \begin{bmatrix}
        \presub{\mathcal{P}_i}{\kappa}_\mathrm{b}\\ \presub{\mathcal{P}_i}{\sigma}_\mathrm{sh}\\ \presub{\mathcal{P}_i}{\sigma}_\mathrm{ax}
    \end{bmatrix} - \begin{bmatrix}
        \kappa_\mathrm{be}^0\\ \sigma_\mathrm{sh}^0\\ \sigma_\mathrm{ax}^0 + \epsilon_i (\phi_i)
    \end{bmatrix} \right ),
\end{equation}
where $\presub{\mathcal{P}_i}{\xi}^0 = \begin{bmatrix}\kappa_\mathrm{be}^0 & \sigma_\mathrm{sh}^0 & \sigma_\mathrm{ax}^0\end{bmatrix}^\mathrm{T}$ denotes the rest strain,
$S_{\mathrm{be},i}(\phi_i)$, $S_{\mathrm{sh},i}(\phi_i)$, $S_{\mathrm{ax},i}(\phi_i)$ are the bending, shear, and axial stiffnesses which are defined as linear functions with respect to the twist angle of the rod $\phi_i$~\cite{good2022expanding, stolzle2023modelling}:
\begin{equation}\small
    S_{\mathrm{be},i}(\phi_i) = \hat{S}_{\mathrm{b}} + C_{\mathrm{S}_\mathrm{b}} \, \phi_{i}^+,
    \quad
    S_{\mathrm{sh},i}(\phi_i) = \hat{S}_{\mathrm{sh}} + C_{\mathrm{S}_\mathrm{sh}} \, \phi_{i}^+,
    \quad 
    S_{\mathrm{ax},i}(\phi_i) = \hat{S}_{\mathrm{ax}} + C_{\mathrm{S}_\mathrm{ax}} \, \phi_{i}^+.
\end{equation}
The coefficient $S_{\mathrm{b},\mathrm{sh}}$ accounts for the elastic coupling between the bending and the shear strain. 
Subsequently, we project the forces into the virtual backbone by premultiplying with $J_\beta^\mathrm{T} = \frac{\partial \beta}{\partial q}^\mathrm{T}$ and then sum the contribution of all rods.
Finally, we group all terms depending on the control input $\phi$ in $\alpha(q,\phi)$ and everything else in $K$.
After modeling the dissipative forces in each \ac{HSA} as $\mathrm{diag}(\zeta_\mathrm{be}, \zeta_{\mathrm{sh}}, \zeta_{\mathrm{ax}}) \, \presub{\mathcal{P}_i}{\dot{\xi}}$, we derive the damping matrix in configuration space as $D = \sum_{i=1}^{2} J_{\beta,i}^\mathrm{T} \, \mathrm{diag}(\zeta_\mathrm{be}, \zeta_{\mathrm{sh}}, \zeta_{\mathrm{ax}}) \, J_{\beta,i} = 2 \, \mathrm{diag}\left ( (\zeta_\mathrm{be} + r_\mathrm{off}^2 \, \zeta_\mathrm{ax} ), \zeta_{\mathrm{sh}}, \zeta_{\mathrm{ax}} \right)$.
We open-source the derivation of the Euler-Lagrangian dynamics and a JAX implementation of a simulator based on them on GitHub\footnote{\url{https://github.com/tud-phi/jax-soft-robot-modelling}}.
We stress that (a) the derived dynamical model is not affine in the control input and (b) the system is underactuated.

\vspace{-1em}\subsection{Control}\label{sub:methodology:control}\vspace{-0.4em}
Our goal is to control the end-effector, which is defined as the distal surface of the platform, to a desired position in Cartesian space $p_\mathrm{ee}^\mathrm{d} \in \mathbb{R}^2$. 
However, the mapping into configuration space is not trivial as we do not know which end-effector orientation $\theta_\mathrm{ee}$ is feasible at steady-state. 
To tackle this challenge, we perform steady-state planning identifying admittable configurations $q^\mathrm{d}$ and matching steady-state actuations $\phi^\mathrm{ss}$, which allow the robot's end-effector to statically remain at $p_\mathrm{ee}^\mathrm{d}$. More details on the used planning procedure can be found in Section~\ref{sub:experiments:steady_state_planning}.

In principle, we can command $\phi = \phi^\mathrm{ss}$ to achieve regulation towards the desired end-effector position.
Nevertheless, we add a feedback controller to compensate for any errors in $\phi^\mathrm{ss}$ caused by unmodelled effects such as hysteresis. Unfortunately, the non-affine actuation $\alpha(q,\phi)$ would complicate the design of such a feedback controller.
Therefore, we perform a first-order Taylor expansion of the actuation forces with respect to $\phi$ resulting in a configuration-dependent actuation matrix $A_{\phi^\mathrm{ss}}(q) = \frac{\partial \alpha}{\partial \phi} \big|_{\phi = \phi_\mathrm{ss}} \in \mathbb{R}^{3 \times 2}$. This allows us to re-write the right side of the \ac{EOM} as $\tau_q = \alpha(q^\mathrm{ss}, \phi^\mathrm{ss}) + A_{\phi^\mathrm{ss}}(q) \, u$ where $u = \phi - \phi^\mathrm{ss}$ is the new control input.
To improve the robustness of the control loop, we compute $u$ with a P-satI-D control law~\cite{pustina2022p}. However, our system is underactuated and in a non-collocated form.
Therefore, we apply a coordinate transformation $h: q \rightarrow \varphi \in \mathbb{R}^3$ recently introduced by Pustina et al.~\cite{pustina2024input} which maps the \ac{EOM} into a form where $\phi$ applies direct forces on the actuated configuration variables. The map is given by {\small $h(q) = \begin{bmatrix}
    \int_0^t \dot{q}^\mathrm{T} A_{\phi^\mathrm{ss}}(q) \mathrm{d}\tau, & \sigma_\mathrm{sh}
\end{bmatrix}^\mathrm{T} = \begin{bmatrix}
    h_1(q), & h_2(q), & \sigma_\mathrm{sh}
\end{bmatrix}^\mathrm{T}$}
with
\vspace{-0.5cm}
\begin{footnotesize}
\begin{multline}\footnotesize
    h_i(q) = 
    C_{\mathrm{S},\mathrm{ax}} \, \frac{h_i}{l^0} \, \Big [ 2 \, \varepsilon_i(\phi^\mathrm{ss}_i) \left ( \pm r_\mathrm{off} \kappa_\mathrm{be} + \sigma_\mathrm{ax} \right ) \mp r_\mathrm{off}^2 \frac{\kappa_\mathrm{be}^2}{2} \pm r_\mathrm{off} \, \sigma_\mathrm{ax}^0 \, \kappa_\mathrm{be} \mp r_\mathrm{off} \, \kappa_\mathrm{be} \, \sigma_\mathrm{ax} + \sigma_\mathrm{ax}^0 \, \sigma_\mathrm{ax}\\ - \frac{\sigma_\mathrm{ax}^2}{2} \Big ] 
    + C_{\mathrm{S},\mathrm{b}} \, \frac{h_i}{l^0} \, \Big [ \kappa_\mathrm{be}^0 \, \kappa_\mathrm{be} - \frac{\kappa_\mathrm{be}^2}{2} \Big ] 
    + C_{\mathrm{S},\mathrm{sh}} \, \frac{h_i}{l^0} \, \Big [\sigma_\mathrm{sh}^0 \, \sigma_\mathrm{sh} - \frac{\sigma_\mathrm{sh}^2}{2} \Big ]
    + \hat{S}_\mathrm{ax} \, \frac{h_i}{l^0} \, C_\varepsilon \Big [ \pm r_\mathrm{off} \, \kappa_\mathrm{be} + \sigma_\mathrm{ax} \Big ].
\end{multline}
\end{footnotesize}
The Jacobian $J_\mathrm{h}(q) = \frac{\partial h}{\partial q}$ is used to formulate the dynamics $M_\varphi \ddot{\varphi} + \eta(\varphi, \dot{\varphi}) + G_\varphi + K_\varphi + D_\varphi \, \dot{\varphi} = J_\mathrm{h}^\mathrm{-T}(q) \, \alpha(q^\mathrm{ss},\phi^\mathrm{ss}) + A_\varphi \, u $ in the collocated variables~\cite{khatib1987unified}, where $A_\varphi^\mathrm{T} = \begin{bmatrix}
    \mathbb{I}^{2} & 0^\mathrm{2 \times 1}
\end{bmatrix}^\mathrm{T}$. In the following, we will denote with the subscript $a$ the first two actuated coordinates $\varphi_\mathrm{a}$.
Finally, the full control law of the \emph{P-satI-D} is given in collocated form as
\begin{equation}\small\label{eq:gravity_compensation_controller}
    \phi = \phi^\mathrm{ss} + K_\mathrm{p} (\varphi_\mathrm{a}^\mathrm{d} - \varphi) - K_\mathrm{d} \dot{\varphi}_\mathrm{a} + K_\mathrm{i} \int_0^t \tanh(\gamma \, ( \varphi_{\mathrm{a},t'}^\mathrm{d}-\varphi_{\mathrm{a},t'})) \: \mathrm{d} t',
\end{equation}
where $K_\mathrm{p}, K_\mathrm{d}, K_\mathrm{i} \in \mathbb{R}^{2 \times 2}$ are the proportional, derivative, and integral gains respectively, and $\gamma \in \mathbb{R}^{2 \times 2}$ horizontally compresses the hyperbolic tangent. While the proposed P-satI-D control law compensates gravity through $\phi^\mathrm{ss}$, we can extend the approach to include gravity cancellation (\emph{P-satI-D + GC}) by evaluating $G_{\varphi,\mathrm{a}}$ at the current configuration:
\begin{equation}\small\label{eq:gravity_cancellation_controller}
    \phi = \phi^\mathrm{ss} - G_{\varphi,\mathrm{a}}(q^\mathrm{d}) + G_{\varphi,\mathrm{a}}(q) + K_\mathrm{p} (\varphi_\mathrm{a}^\mathrm{d} - \varphi) - K_\mathrm{d} \dot{\varphi}_\mathrm{a} + K_\mathrm{i} \int_0^t \tanh(\gamma \, ( \varphi_{\mathrm{a},t'}^\mathrm{d}-\varphi_{\mathrm{a},t'})) \: \mathrm{d} t'.
\end{equation}
The implementation of all control laws is available on GitHub\footnote{\url{https://github.com/tud-phi/hsa-planar-control}}.

\begin{figure}[ht]
    \centering
    \subfigure[FPU: End-effector pose]{\includegraphics[width=0.48\columnwidth, trim={10, 10, 10, 10}]{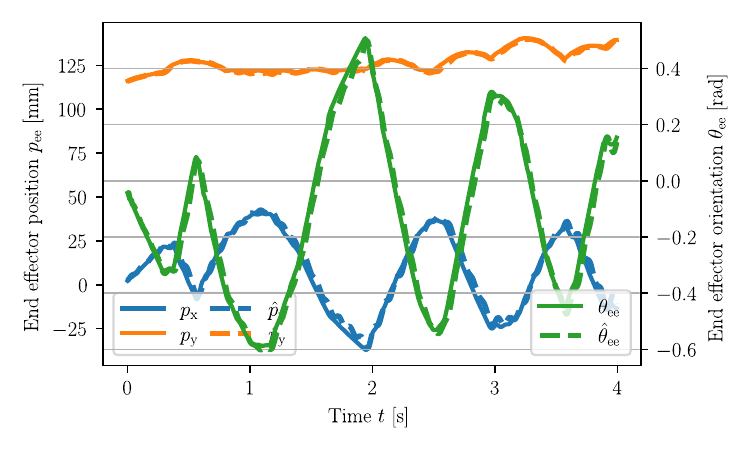}\label{fig:model_verification:fpu:chiee}}
    \subfigure[FPU: Configuration]{\includegraphics[width=0.48\columnwidth, trim={10, 10, 10, 10}]{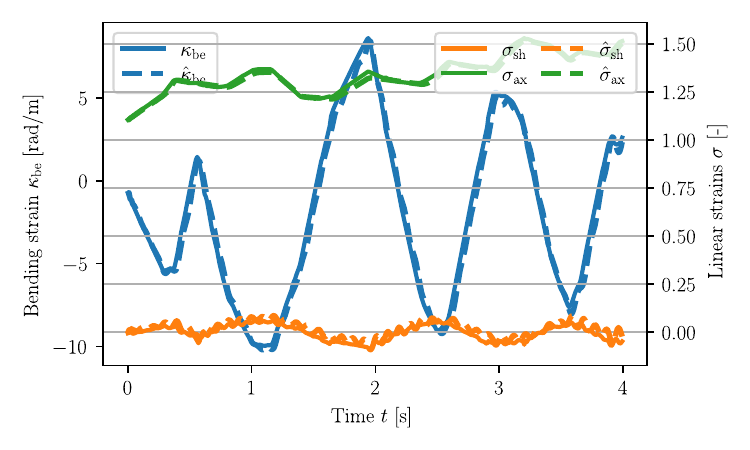}\label{fig:model_verification:fpu:q}}\\
    \vspace{-0.1cm}
    \subfigure[EPU: End-effector pose]{\includegraphics[width=0.48\columnwidth, trim={10, 10, 10, 10}]{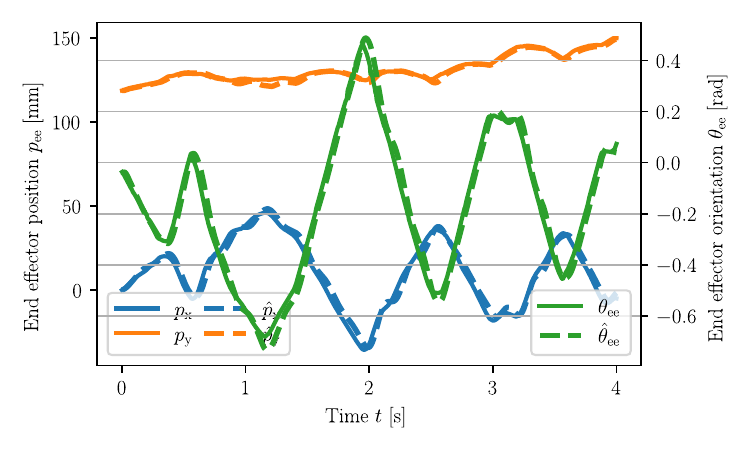}\label{fig:model_verification:epu:chiee}}
    \subfigure[EPU: Configuration]{\includegraphics[width=0.48\columnwidth, trim={10, 10, 10, 10}]{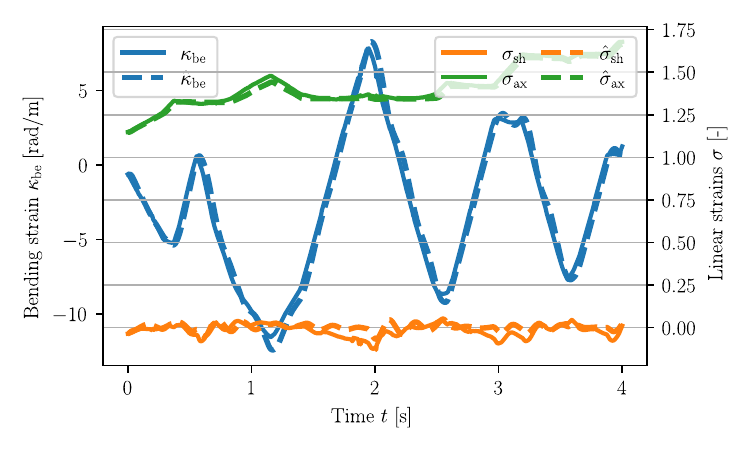}\label{fig:model_verification:epu:q}}
    \vspace{-0.2cm}
    \caption{Verification of the system model and the identified system parameters on an unseen trajectory with the HSA being randomly actuated through a GBN sequence: the solid line denotes the actual trajectory. In contrast, the dashed line visualizes the trajectory simulated with the system model. We report results for both FPU and EPU-based \acp{HSA}.}\label{fig:model_verification}
\end{figure}

\vspace{-1em}\section{Experimental validation}

\subsection{Experimental setup}\vspace{-0.4em}
We evaluate the system model and our proposed control approach on a robot consisting of four HSA rods.
The material choice of the \ac{HSA} is crucial and has a significant influence on the resulting mechanical characteristics of the robot  (e.g., blocked force, holding torque, bending stiffness, etc.)~\cite{truby2021recipe}. Furthermore, specific material requirements are dictated by the nature of the design of the \ac{HSA} rod. The structure of the metamaterial is made of struts connected by living hinges. These living hinges must be thin, flexible, and accommodate high strains~\cite{truby2021recipe}.
Therefore, we decided to 3D-print the \acp{HSA} via digital projection lithography either from the photopolymer resin Carbon FPU 50 (stiffer) or the elastomeric polyurethane EPU 40 resin (softer).

Each \ac{HSA} rod is actuated by a Dynamixel MX-28 servo motor. The Dynamixel motors are set to use position control mode. 
The robot is mounted platform-down on a cage with an Optitrack motion capture system, which measures the SE(3) pose of the platform at \SI{200}{Hz}.
Our algorithms run within a ROS2 framework\footnote{\url{https://github.com/tud-phi/ros2-hsa}}. 
The pose measurements are first projected into the plane of actuation and serve as an input to the closed-form inverse kinematics introduced in \eqref{eq:kinematics}. 
We use a Savitzky-Golay filter with a window duration of $\SI{0.1}{s}$ to numerically differentiate $\chi_\mathrm{ee}(t)$, $q(t)$ and gather with that $\dot{\chi}_\mathrm{ee}(t)$ and $\dot{q}(t)$.

\vspace{-0.2em}\subsection{System identification}\vspace{-0.4em}
Next, we strive to identify the parameters used in our dynamic model.
We assume the robot's geometric and mass density properties to be known or easily measurable. 
As knowledge about the damping coefficients is not required by the control law, only the experimental identification of elongation and stiffness characteristics remains.
For this, we measure the response of the system to step and staircase actuation sequences. Afterward, the parameters are regressed using least squares. 
For the FPU-based robot, we identify $C_\varepsilon^\mathrm{FPU}=\SI{0.0098}{m \per rad}$, $S_\mathrm{be}^\mathrm{FPU} = 0.00057 \cdot 10^{-5} - 9.7 \cdot 10^{-6} \, \frac{\phi_i^+}{l^0} \si{Nm^2}$, $S_\mathrm{sh}^\mathrm{FPU} = 0.591 - 0.00048 \, \frac{\phi_i^+}{l^0} \si{N}$, $S_\mathrm{ax}^\mathrm{FPU} = 5.665 + 0.0151 \, \frac{\phi_i^+}{l^0} \si{N}$, and $S_\mathrm{b,sh}^\mathrm{FPU} = 4.48 \cdot 10^{-3} \si{Nm \per rad}$ where $l^0 = \SI{0.059}{m}$. 
Furthermore, we regress $C_\varepsilon^\mathrm{EPU}=\SI{0.0079}{m \per rad}$, $S_\mathrm{be}^\mathrm{EPU} = -2.5 \cdot 10^{-5} + 3.9 \cdot 10^{-7} \, \frac{\phi_i^+}{l^0} \si{Nm^2}$, $S_\mathrm{sh}^\mathrm{EPU} = 0.0428 - 0.0029 \, \frac{\phi_i^+}{l^0} \si{N}$, $S_\mathrm{ax}^\mathrm{EPU} = 0.0 + 0.0098 \, \frac{\phi_i^+}{l^0} \si{N}$, and $S_\mathrm{b,sh}^\mathrm{EPU} = -\SI{0.0005}{Nm \per rad}$ for the EPU \acp{HSA} which have the same length as the FPU \acp{HSA}.
Finally, we identify the axial rest strain $\sigma_\mathrm{ax}^0$ before the start of each experiment.
We notice that the EPU-based HSA robot is approximately one order of magnitude more flexible than the FPU-based robot.

\begin{figure}[ht]
    \vspace{-0.5cm}
    \centering
    \subfigure[End-effector position]{\includegraphics[width=0.49\columnwidth, trim={10, 10, 10, 10}]{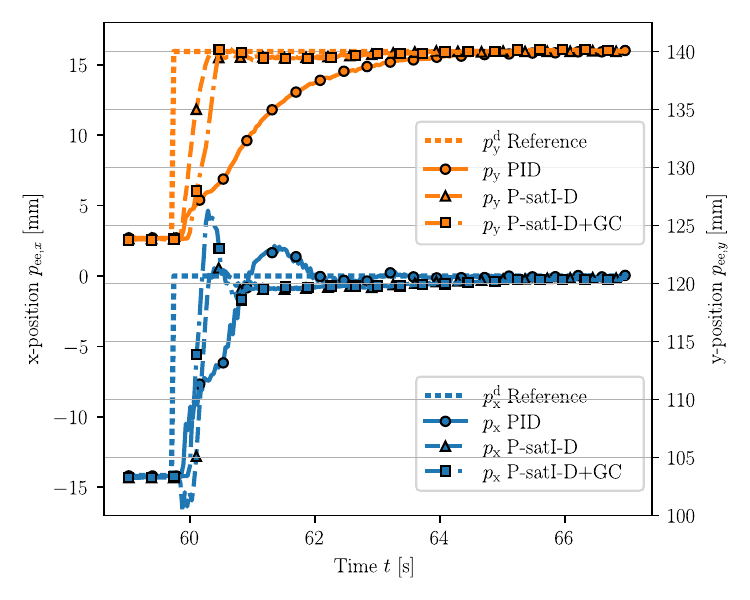}\label{fig:experimental_results:fpu:step_response:pee}}
    \subfigure[Configuration]{\includegraphics[width=0.49\columnwidth, trim={10, 10, 10, 10}]{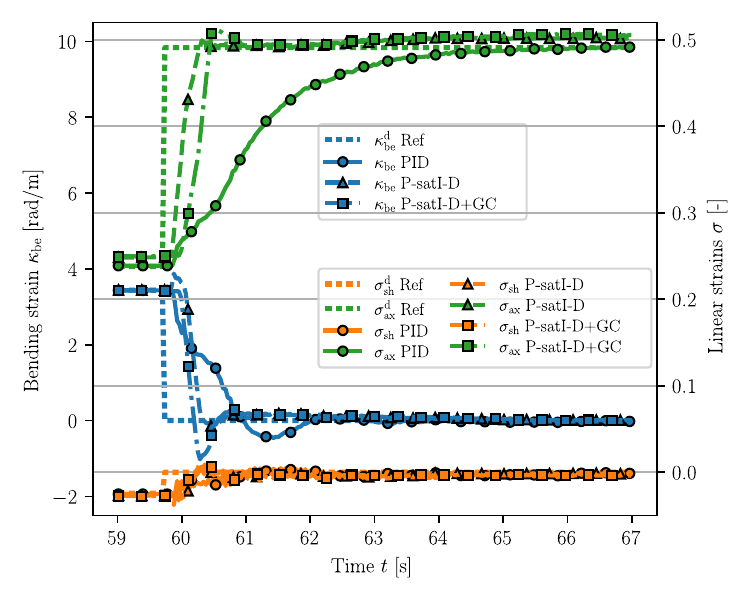}\label{fig:experimental_results:fpu:step_response:q}}
    \vspace{-0.2cm}
    \caption{Step response of the \emph{baseline PID}, \emph{P-satI-D} (with gravity compensation), and \emph{P-satI-D + GC} (with gravity cancellation) controllers on an FPU-based HSA robot.}\label{fig:experimental_results:fpu:step_response}
\end{figure}

\vspace{-1em}\subsection{Model verification}\label{sub:experiments:model_verification}\vspace{-0.4em}
We verify the accuracy of the proposed system model and the identified parameters on trajectories unseen during system identification. We generate the trajectories by actuating the robot with a \ac{GBN}~\cite{tulleken1990generalized} sequence with a settling time of \SI{0.5}{s} and at each time step $k$ randomly sample $\phi(k) \sim \mathcal{U}(0, \phi_\mathrm{max})$.
We simulate the model evolution with a Dormand-Prince 5(4) integrator and a time step of \SI{0.1}{ms}.
Fig.~\ref{fig:model_verification:fpu:chiee} shows the model exhibiting excellent accuracy for representing the behavior of FPU-based \ac{HSA} robots.
We observe more significant errors in the shear estimate for EPU-based \ac{HSA} robots in Fig.\ref{fig:model_verification:epu:q}. Specifically, the \ac{CS} model no longer seems sufficient for capturing the robot's shape, particularly for larger bending angles. Therefore, we suggest for future work to employ kinematic models with more \ac{DOF} such as \ac{PCS} as proposed, for example, in \cite{stolzle2023modelling}.

\begin{figure}[t]
    \centering
    \subfigure[End-effector position]{\includegraphics[width=0.32\columnwidth, trim={10, 10, 5, 10}]{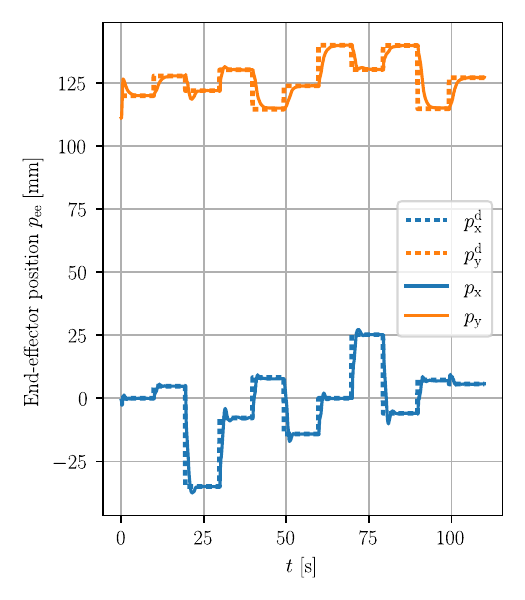}\label{fig:experimental_results:fpu:baseline_pid:pee}}
    \subfigure[Configuration]{\includegraphics[width=0.32\columnwidth, trim={10, 10, 10, 10}]{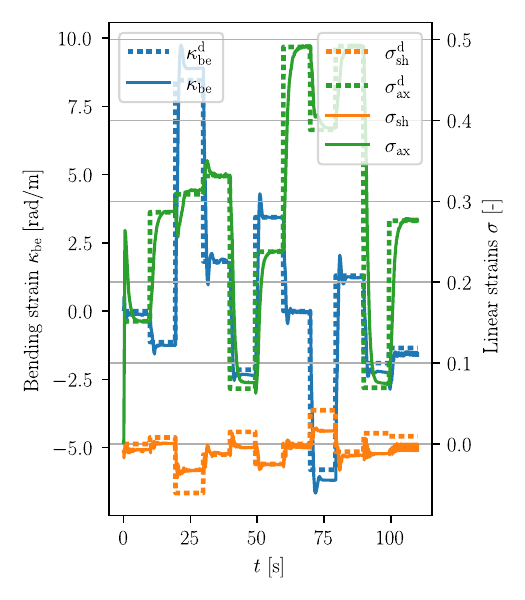}\label{fig:experimental_results:fpu:baseline_pid:q}}
    \subfigure[Control input]{\includegraphics[width=0.32\columnwidth, trim={10, 10, 10, 10}]{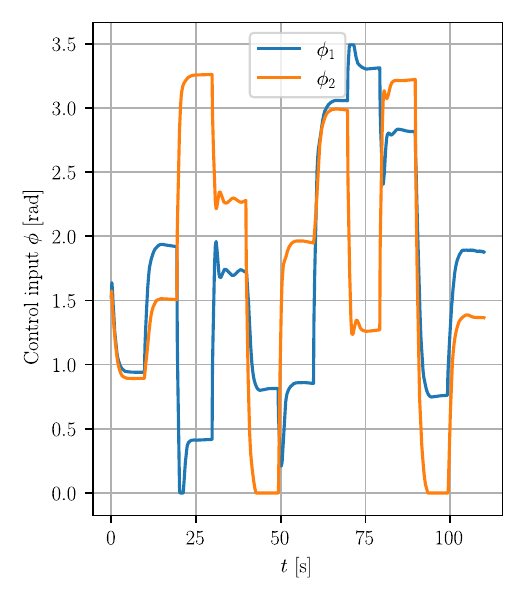}\label{fig:experimental_results:fpu:baseline_pid:phi}}
    \vspace{-0.4cm}
    \caption{Experimental results for tracking a reference trajectory of eleven step functions with the baseline PID controller on an FPU-based HSA robot. \textbf{Panel (a):} End-effector position with the dotted and solid lines denoting the task-space reference and actual position, respectively.
    \textbf{Panel (b):} The planned (dotted) and the actual (solid) configuration. 
    \textbf{Panel (c):} The planned (dotted) and the actual (solid) actuation coordinates of the collocated system. 
    \textbf{Panel(d):} The saturated planar control inputs are visualized with solid lines, and the computed steady-state actuation with dotted lines.}\label{fig:experimental_results:fpu:baseline_pid}
\end{figure}

\subsection{Steady-state planning}\label{sub:experiments:steady_state_planning}\vspace{-0.4em}
Our approach, as detailed in Section~\ref{sub:methodology:control}, requires us for a given desired end-effector position $p_\mathrm{ee}^\mathrm{d}$ to identify a statically-feasible configuration $q^\mathrm{d}$ with the matching steady-state actuation $\phi^\mathrm{ss}$.

We perform online static inversion to identify admittable desired configurations $q^\mathrm{d}$ and matching steady-state control inputs $\phi^\mathrm{ss}$ during our experiments involving the FPU \ac{HSA} robots. First, we substitute the inverse kinematics $\varrho_\mathrm{ee}(\chi_\mathrm{ee})$ into the static \ac{EOM}. Then, we find the roots of the equation $G\circ\varrho_\mathrm{ee}(\chi_\mathrm{ee}^\mathrm{d}) + K\circ\varrho_\mathrm{ee}(\chi_\mathrm{ee}^\mathrm{d})-\alpha(\varrho_\mathrm{ee}(\chi_\mathrm{ee}^\mathrm{d}), \phi_\mathrm{ss})$ with respect to $(\theta_\mathrm{ee},\phi_1, \phi_2)$ using nonlinear least-squares while enforcing constraints on the sign of $\phi$. We solve this optimization problem with projected gradient descent.

In contrast, the static inversion optimization problem is not well-behaved for the identified EPU system parameters. Instead, we rely on rolling out the dynamics over a duration $t_\mathrm{ss}$ to steady-state and then optimize the steady-state input $\phi^\mathrm{ss}$ such that the final end-effector error $\lVert p_\mathrm{ee}^\mathrm{d} - p_\mathrm{ee}^\mathrm{ss} \rVert$ is as small as possible. We formalize this optimization problem in a least-squares fashion
\begin{equation}\small\label{eq:steady_state_rollout_optim_problem}
\begin{aligned}
    \phi^\mathrm{ss} = \argmin_\mathrm{\phi} \quad & \frac{1}{2} \, \lVert p_\mathrm{ee}^\mathrm{d} - p_\mathrm{ee}^\mathrm{ss}(\phi) \rVert_2^2,\\
    \textrm{s.t.} \quad & x^\mathrm{ss} = x(t_0) + \int_{t_0}^{t_\mathrm{ss}} f(x(t), \phi) \, \mathrm{d}t, \quad \chi_\mathrm{ee}^\mathrm{ss} = \begin{bmatrix}
        p_\mathrm{ee}^\mathrm{ss}\\
        \theta_\mathrm{ee}^\mathrm{ss}
    \end{bmatrix} = \pi_\mathrm{ee}(q^\mathrm{ss}),\\
\end{aligned}
\end{equation}
where $\dot{x}(t) = f(x(t), \phi)$ are the nonlinear state-space dynamics based on the \ac{EOM} derived in Section~\ref{sub:methodology:dynamics} and $\phi \in \mathbb{R}^2$ is constant in time. We solve \eqref{eq:steady_state_rollout_optim_problem} online using the Levenberg-Marquardt algorithm. Finally, we choose $q^d = q^\mathrm{ss}$ and $\chi_\mathrm{ee}^\mathrm{d} = \pi_\mathrm{ee}(q^d)$.

\vspace{-1em}\subsection{Closed-loop control}\vspace{-0.4em}
Next, we implement the closed-loop control strategy laid out in Section~\ref{sub:methodology:control}.
After evaluating the control law at a rate of \SI{40}{Hz} and saturating the control inputs to the ranges $[0, 3.40] \, \si{rad}$ for FPU and $[0, 4.71] \, \si{rad}$ for EPU, respectively, we map $\phi \in \mathbb{R}^2$ to desired positions of the four motors. For this, we consider the handedness of the \acp{HSA} and apply the same actuation magnitude to both rods on the same side of the virtual backbone.
After tuning the gains for the feedback part of the model-based control laws in \eqref{eq:gravity_compensation_controller} and \eqref{eq:gravity_cancellation_controller}, we select $K_\mathrm{p} = \mathrm{diag}(0.3, 0.3)$, $K_\mathrm{i} = \mathrm{diag}(0.05, 0.05) \, \si{1 \per s}$, $K_\mathrm{d} = \mathrm{diag}(0.01, 0.01) \, \si{s}$, and $\gamma = \mathrm{diag}(100, 100)$. 
Furthermore, we report the performance of a model-free PID controller as a baseline. Here, the control input in task-space is given by $u_\mathrm{ts} = \begin{bmatrix}u_\mathrm{ts,x} & u_\mathrm{ts,y}\end{bmatrix}^\mathrm{T} = K_\mathrm{p}^\mathrm{PID} \, (p_\mathrm{ee}^\mathrm{d}-p_\mathrm{ee}) - K_\mathrm{d}^\mathrm{PID} \, \dot{p}_\mathrm{ee} + K_\mathrm{i}^\mathrm{PID} \int_0^t p_{\mathrm{ee},t'}^\mathrm{d} - p_{\mathrm{ee},t'} \: \mathrm{d}t'$, which is then mapped to the actuation via $\phi = \begin{bmatrix}
    u_\mathrm{ts,x}+u_\mathrm{ts,y}, & -u_\mathrm{ts,x}+u_\mathrm{ts,y}
\end{bmatrix}^\mathrm{T}$.
Here, we select $K_\mathrm{p}^\mathrm{PID} = \mathrm{diag}(10, 10) \, \si{rad \per m}$, $ K_\mathrm{i}^\mathrm{PID} = \mathrm{diag}(110, 110) \, \si{rad \per \meter \per \second}$, and $ K_\mathrm{d}^\mathrm{PID} = \mathrm{diag}(0.25, 0.25) \, \si{rad \, \second \per \meter}$.
\\

\begin{figure}[ht]
    \vspace{-0.5cm}
    \centering
    \subfigure[End-effector position]{\includegraphics[width=0.49\columnwidth, trim={5, 10, 5, 5}]{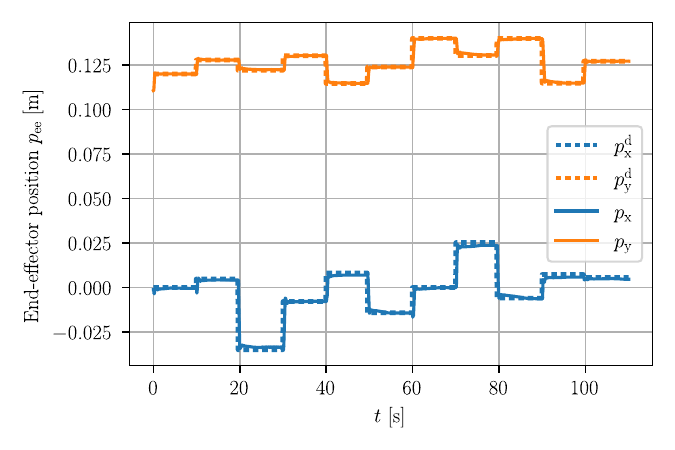}\label{fig:experimental_results:fpu:p_sati_d:pee}}
    \subfigure[Configuration]{\includegraphics[width=0.49\columnwidth, trim={5, 10, 5, 5}]{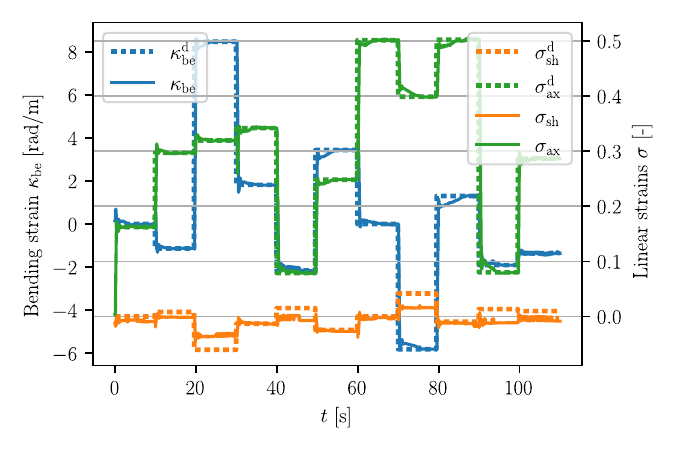}\label{fig:experimental_results:fpu:p_sati_d:q}}\\
    \vspace{-0.3cm}
    \subfigure[Actuation coordinates]{\includegraphics[width=0.49\columnwidth, trim={5, 10, 5, 5}]{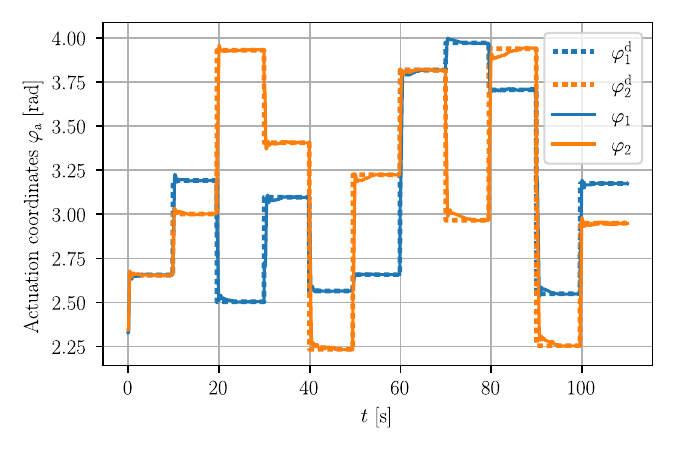}\label{fig:experimental_results:fpu:p_sati_d:varphi}}
    \subfigure[Control input]{\includegraphics[width=0.49\columnwidth, trim={5, 10, 5, 5}]{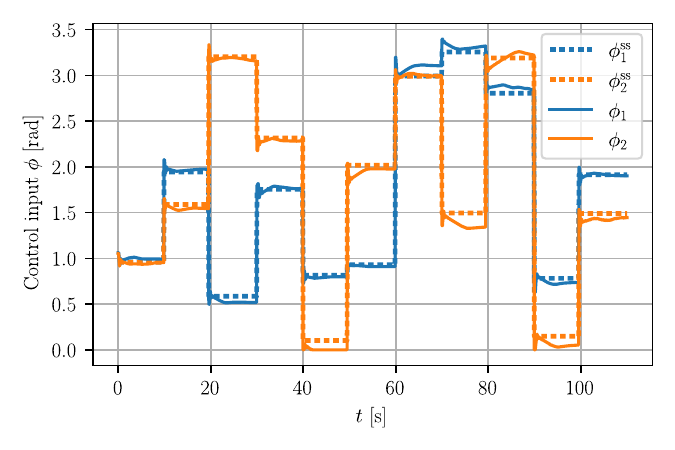}\label{fig:experimental_results:fpu:p_sati_d:phi}}\\
    \vspace{-0.4cm}
    \caption{Experimental results for tracking a reference trajectory of eleven step functions with the P-satI-D controller on an FPU-based HSA robot. \textbf{Panel (a):} End-effector position with the dotted and solid lines denoting the task-space reference and actual position, respectively.
    \textbf{Panel (b):} The planned (dotted) and the actual (solid) configuration. 
    \textbf{Panel (c):} The planned (dotted) and the actual (solid) actuation coordinates of the collocated system. 
    \textbf{Panel(d):} The saturated planar control inputs are visualized with solid lines, and the computed steady-state actuation with dotted lines.}\label{fig:experimental_results:fpu:p_sati_d}
\end{figure}

\noindent\textbf{Evaluation:}
We define a reference trajectory $p_\mathrm{ee}^\mathrm{d}(k), k \in \{ 1, \dots, n_k \}$ with a duration of \SI{110}{s} and consisting of eleven step functions as the reference trajectory.
We report the \ac{RMSE} metric $\sqrt{\sum_{k=1}^{n_k} \frac{\lVert p_\mathrm{ee}^\mathrm{d}(k) - p_\mathrm{ee}(k) \rVert_2^2}{n_k}}$ for assessing the control performance, where $p_\mathrm{ee}(k)$ is the actual trajectory of the end-effector.\\

\begin{figure}[ht]
    \vspace{-0.5cm}
    \centering
    \subfigure[Star]{\includegraphics[width=0.32\columnwidth, trim={8, 8, 8, 8}]{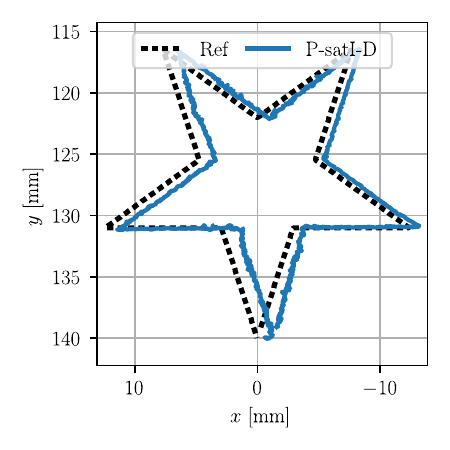}\label{fig:experimental_results:fpu:star}}
    \subfigure[TUD flame]{\includegraphics[width=0.32\columnwidth, trim={8, 8, 8, 8}]{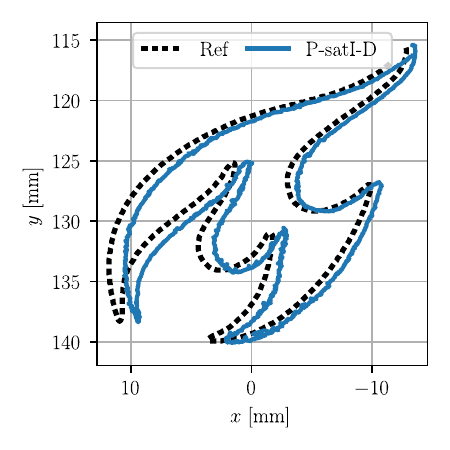}\label{fig:experimental_results:fpu:tud_flame}}
    \subfigure[MIT-CSAIL]{\includegraphics[width=0.32\columnwidth, trim={8, 8, 8, 8}]{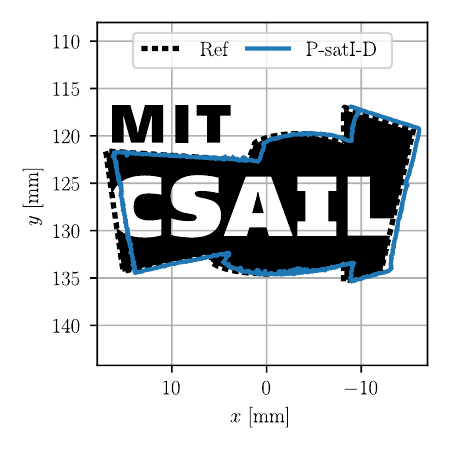}\label{fig:experimental_results:fpu:mit_csail}}\\
    \vspace{-0.2cm}
    \subfigure[Bat trajectories]{\includegraphics[width=0.7\columnwidth, trim={10, 10, 10, 10}]{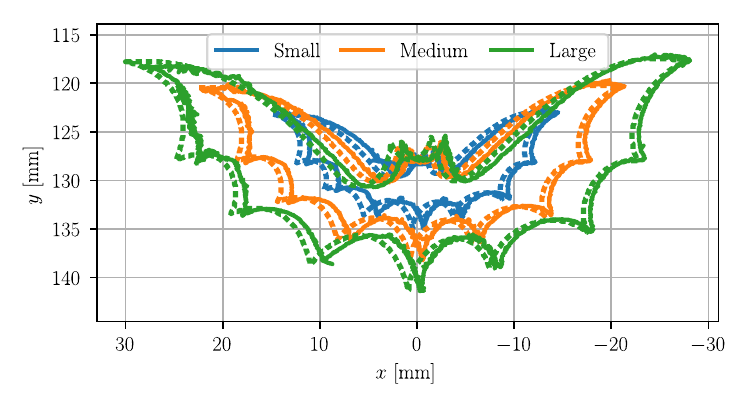}\label{fig:experimental_results:fpu:bats}}
    \vspace{-0.4cm}
    \caption{Cartesian evolution of the proposed P-sat-D controller (solid lines) tracking various continuous reference trajectories (dotted lines) on the FPU robot.
    }\label{fig:experimental_results:fpu:task_space_trajectories}
\end{figure}

\noindent\textbf{Control of an FPU-based HSA robot:}
The \emph{baseline PID} achieves an \ac{RMSE} of \SI{5.86}{mm} with respect to the reference trajectory. The \emph{P-satI-D} based on \eqref{eq:gravity_compensation_controller} (with gravity compensation) exhibits an RMSE of \SI{4.17}{mm}. Similarly, the \emph{P-satI-D + GC} based on \eqref{eq:gravity_cancellation_controller} (with gravity cancellation) displays an RMSE of \SI{4.13}{mm}.
We present a comparison of the three different controllers for a step response in Fig.~\ref{fig:experimental_results:fpu:step_response} and plot the entire trajectories of the \emph{baseline PID} and the \emph{P-satI-D} in Figures~\ref{fig:experimental_results:fpu:baseline_pid} and \ref{fig:experimental_results:fpu:p_sati_d}, respectively.
Additionally, we discretize various continuous reference trajectories into setpoints: 
star trajectory ($873$ setpoints and duration of \SI{109}{s}), the flame of the TU Delft logo ($680$ setpoints and duration of \SI{85}{s}), the contour of the MIT-CSAIL logo ($1046$ setpoints and duration of \SI{131}{s}), and the outline of a bat at three different sizes ($1510$ setpoints and \SI{189}{s} duration).
The resulting Cartesian evolutions of the \emph{P-satI-D} controller tracking these continuous references are displayed in Fig.~\ref{fig:experimental_results:fpu:task_space_trajectories}.

\begin{figure}[hb]
    \centering
    \subfigure[t=\SI{0}{s}]{\includegraphics[width=0.192\columnwidth]{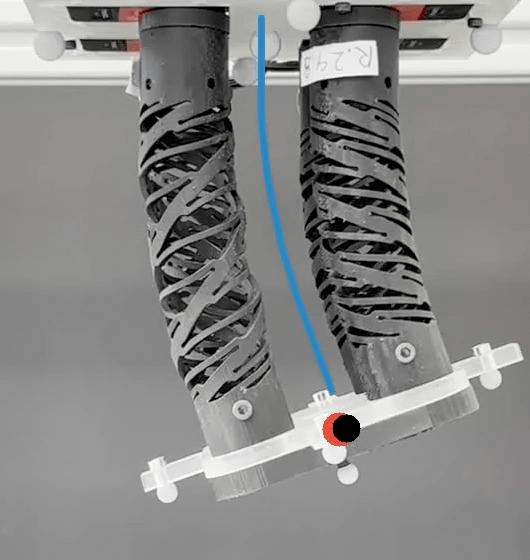}}
    \subfigure[t=\SI{47}{s}]{\includegraphics[width=0.192\columnwidth]{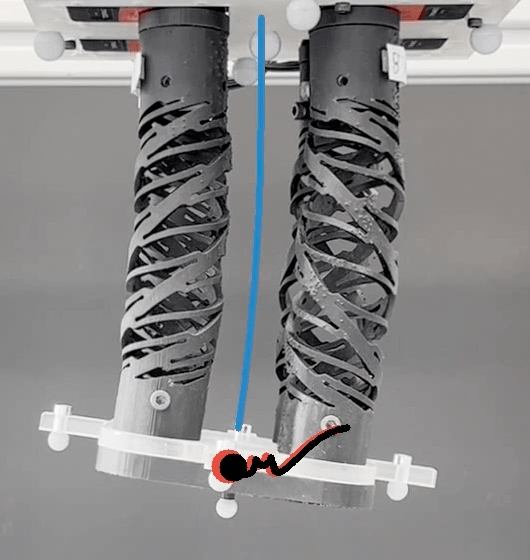}}
    \subfigure[t=\SI{94}{s}]{\includegraphics[width=0.192\columnwidth]{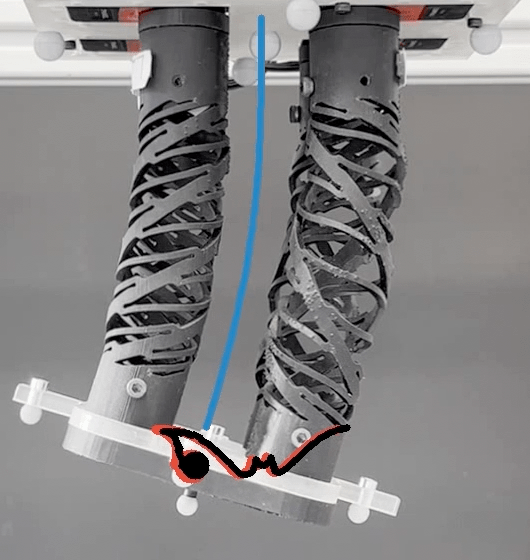}}
    \subfigure[t=\SI{141}{s}]{\includegraphics[width=0.192\columnwidth]{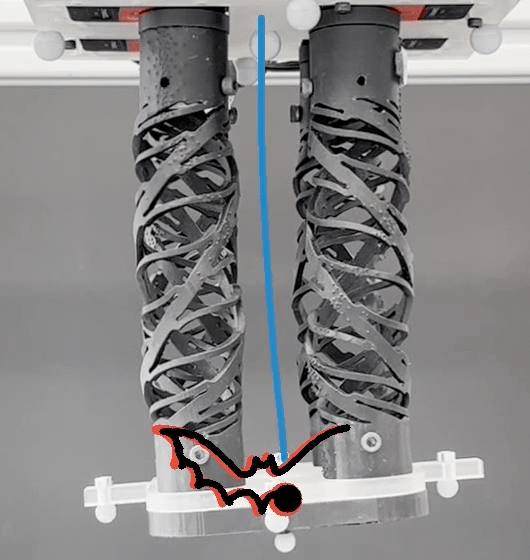}}
    \subfigure[t=\SI{188}{s}]{\includegraphics[width=0.192\columnwidth]{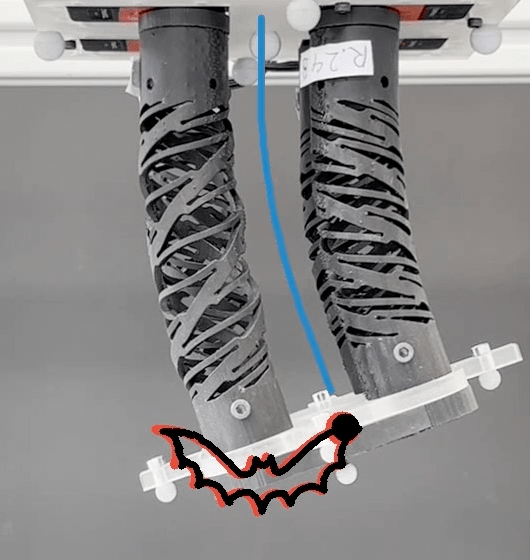}}
    \vspace{-0.2cm}
    \caption{Sequence of stills for the large bat trajectory performed with the P-satD controller on the FPU robot. The red and black dots visualize the desired and current end-effector positions, respectively. The past trajectory is plotted in red (reference) and black (actual). The blue line renders the shape of the virtual backbone.
    }\label{fig:experimental_results:fpu:sequence_of_stills:bat}
\end{figure}

The step response in Fig.~\ref{fig:experimental_results:fpu:step_response} shows how the two model-based controllers \emph{P-satI-D} and \emph{P-satI-D + GC} can leverage the planned $\phi^\mathrm{ss}$ and $q^\mathrm{d}$ to achieve a fast response time of roughly \SI{1.2}{s}. In contrast, the baseline PID needs to wait for the integral error to build up and thus has a much slower response time of approximately \SI{4.2}{s}. Furthermore, overshooting caused by the baseline PID is usually more extensive than that caused by the model-based controllers.
We conclude that \emph{P-satI-D} (gravity compensation) and \emph{P-satI-D + GC} (gravity cancellation) exhibit quite similar behavior. Sometimes, \emph{P-satI-D} exhibits undershooting at the beginning of the transient and \emph{P-satI-D + GC} overshooting towards the end of the transient (see Fig.~\ref{fig:experimental_results:fpu:step_response:pee}).

\begin{figure}[ht]
    \vspace{-0.2cm}
    \centering
    \subfigure[End-effector position]{\includegraphics[width=0.47\columnwidth, trim={10, 10, 10, 10}]{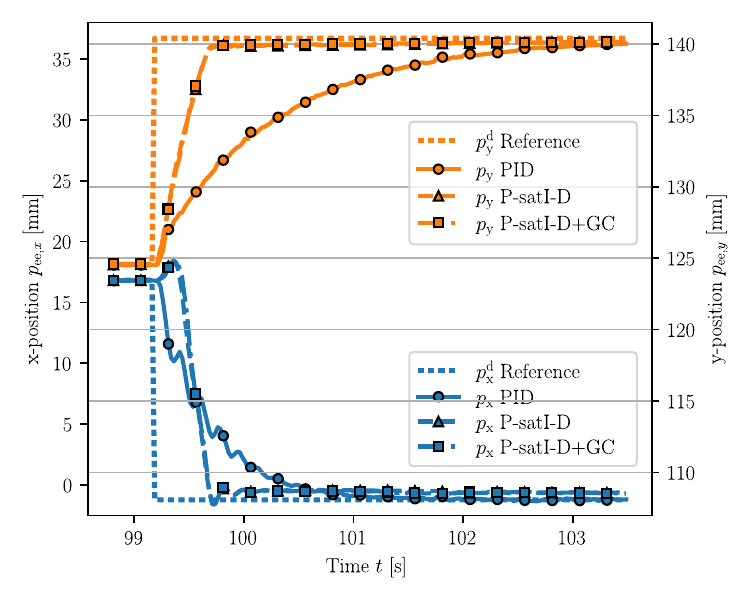}\label{fig:experimental_results:epu:step_response:pee}}
    \subfigure[Configuration]{\includegraphics[width=0.47\columnwidth, trim={10, 10, 10, 10}]{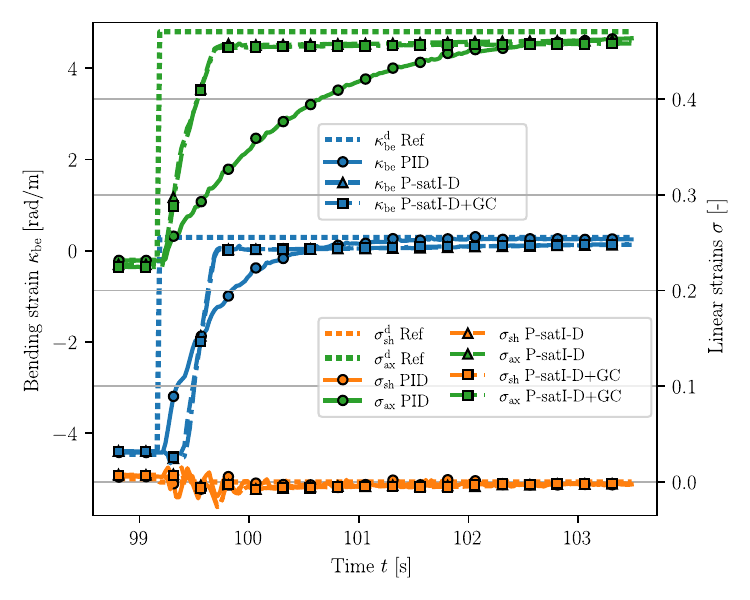}\label{fig:experimental_results:epu:step_response:q}}
    \vspace{-0.2cm}
    \caption{Step responses of the \emph{baseline PID}, \emph{P-satI-D} (with gravity compensation), and \emph{P-satI-D + GC} (with gravity cancellation) controllers on an EPU-based HSA robot.}\label{fig:experimental_results:epu:step_response}
\end{figure}

\begin{figure}[ht]
    \vspace{-0.5cm}
    \centering
    \subfigure[End-effector position]{\includegraphics[width=0.49\columnwidth, trim={5, 10, 5, 5}]{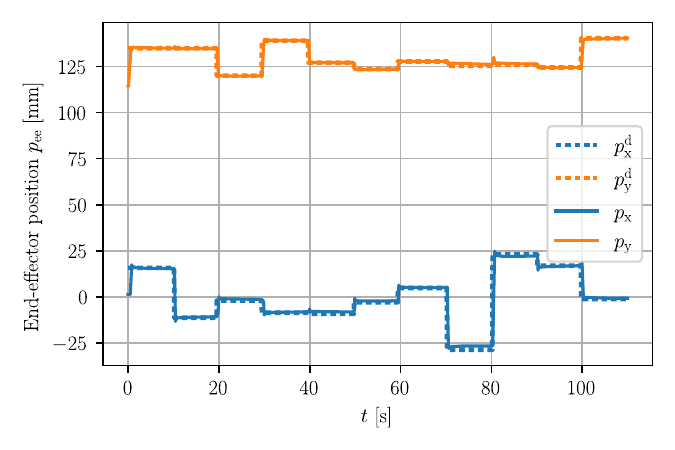}\label{fig:experimental_results:epu:p_sati_d:pee}}
    \subfigure[Configuration]{\includegraphics[width=0.49\columnwidth, trim={5, 10, 5, 5}]{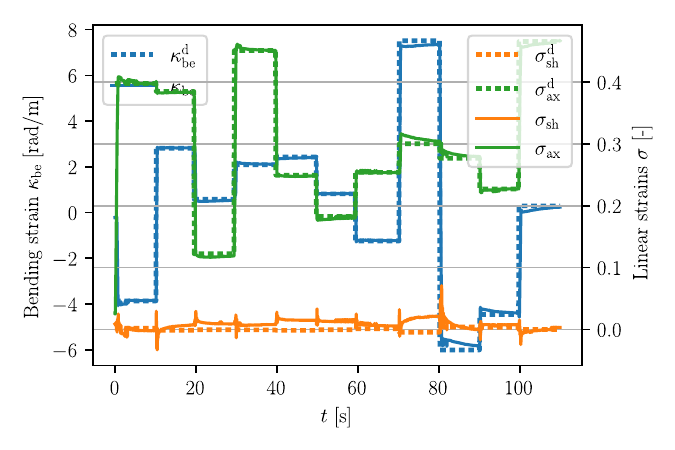}\label{fig:experimental_results:epu:p_sati_d:q}}\\
    \vspace{-0.3cm}
    \subfigure[Actuation coordinates]{\includegraphics[width=0.49\columnwidth, trim={5, 10, 5, 5}]{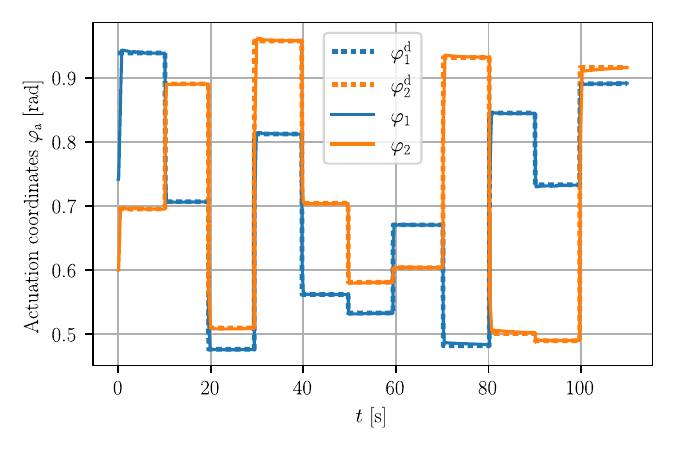}\label{fig:experimental_results:epu:p_sati_d:varphi}}
    \subfigure[Control input]{\includegraphics[width=0.49\columnwidth, trim={5, 10, 5, 5}]{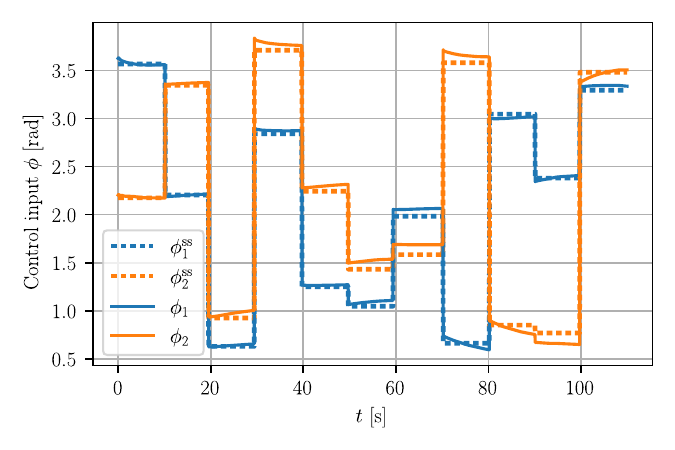}\label{fig:experimental_results:epu:p_sati_d:phi}}
    \vspace{-0.4cm}
    \caption{Experimental results for tracking a reference trajectory of eleven step functions with the P-satI-D controller on an EPU-based HSA robot. \textbf{Panel (a):} End-effector position with the dotted and solid lines denoting the task-space reference and actual position, respectively.
    \textbf{Panel (b):} The planned (dotted) and the actual (solid) configuration. 
    \textbf{Panel (c):} The planned (dotted) and the actual (solid) actuation coordinates of the collocated system. 
    \textbf{Panel(d):} The saturated planar control inputs are visualized with solid lines, and the computed steady-state actuation with dotted lines.}\label{fig:experimental_results:epu:p_sati_d}
\end{figure}

\noindent \textbf{Control of an EPU-based HSA robot:}
Tracking the reference trajectory of eleven step functions with an EPU-based robot, the \emph{baseline PID} controller has an \ac{RMSE} of \SI{4.40}{mm}. The \emph{P-satI-D} (with gravity compensation) can able to achieve an \ac{RMSE} of \SI{3.63}{mm}. The \emph{P-satI-D + GC} controller exhibits similar performance(\ac{RMSE} of \SI{3.71}{mm}).
We visualize the step response of all three controllers in Fig.~\ref{fig:experimental_results:epu:step_response} and the entire trajectory of the \emph{P-satI-D} controller in Fig.~\ref{fig:experimental_results:epu:p_sati_d}.

Again, we notice that the response time of the model-based controllers (\SI{0.54}{s}) is much shorter than the response time of the baseline PID (\SI{3.84}{s}). Furthermore, the importance of a model-based control law is motivated by the oscillations in the transient of the baseline PID (see x-coordinate in Fig.~\ref{fig:experimental_results:epu:step_response:pee}).
The steady-state error for the model-based controllers on the EPU material is slightly higher compared to the FPU material, as seen in Figures \ref{fig:experimental_results:epu:step_response:pee} \& \ref{fig:experimental_results:epu:p_sati_d:pee}. In Section~\ref{sub:experiments:model_verification}, we noticed that the shear model doesn't fully capture the actual system behavior. This then results in an error in the planned desired configuration $q^\mathrm{d}$, which the controller is not able to resolve because of the underactuation of the robot (see Fig.~\ref{fig:experimental_results:epu:p_sati_d:q}).

\section{Experimental insights}
This work shows effective, model-based regulation with planar \ac{HSA} robots. The conducted experiments gave us deep insights into the particular characteristics of \acp{HSA} and how well our model is able to capture them. We see excellent agreement for predicting the dynamical behavior of \ac{HSA} robots made of FPU material.
For EPU-based \acp{HSA} robots, we observe that the model does not fully capture the shear dynamics.

The excellent agreement of the model with the actual system behavior enables our model-based controllers to perform very well at the task of setpoint regulation.
For the model-based controllers, any mismatch between the dynamic model and the actual system (as analyzed in Section~\ref{sub:experiments:model_verification}) has two impacts: (i) the steady-state planning provides us with a desired configuration $q^\mathrm{d}$ which the underactuated robot cannot achieve. This then, in turn, causes a small steady-state error in the end-effector position as seen for the manual setpoints in Fig.~\ref{fig:experimental_results:fpu:p_sati_d:pee}
and for the continuous references in Fig.\ref{fig:experimental_results:fpu:task_space_trajectories}. This steady-state error is absent in the baseline PID as its integral term acts directly in task space. We suggest that future work include an integral term directly on the end-effector position to remove the remaining steady-state error of the model-based controller. Secondly, as (ii), model errors will lead to an offset in the planned steady-state actuation $\phi^\mathrm{ss}$. Therefore, applying a constant $\phi^\mathrm{ss}$ will not move the robot exactly to $p_\mathrm{ee}^\mathrm{d}$. As shown in Fig.~\ref{fig:experimental_results:fpu:p_sati_d:phi}, the P-satI-D feedback term can compensate for this effect through its proportional and integral terms applied in the collocated variables.


\vspace{-1em}\section{Acknowledgements}\vspace{-0.5em}
The work by Maximilian Stölzle was supported under the European Union’s
Horizon Europe Program from Project EMERGE - Grant Agreement No. 101070918. The authors thank Pietro Pustina for the valuable insights into a coordinate transformation into collocated variables, control of underactuated soft robots, and his help revising the manuscript.

\vspace{-0.4cm}
%
%

\bibliographystyle{styles/bibtex/spmpsci}
\bibliography{root}

\end{document}